%% file: main.tex
\documentclass[a4paper, 10pt, conference]{ieeeconf}
\IEEEoverridecommandlockouts
\usepackage{graphicx}
\usepackage{setspace}          
\usepackage{url}
\usepackage{amsmath,bm}
\usepackage{upgreek}
\usepackage{algorithmic}
\usepackage{amssymb} 
\usepackage{mathtools}
\usepackage{xcolor} 
\usepackage{siunitx}
\usepackage{eucal}
\usepackage{systeme}
\usepackage[noadjust]{cite}
\usepackage{subfig}
\usepackage{cuted}
\usepackage{array}
\usepackage{verbatim}
\usepackage{hyperref}
\usepackage{amsfonts}
\usepackage{amssymb}
\usepackage{microtype}
\usepackage{caption}
\usepackage{glossaries}
\usepackage{graphicx}
\usepackage{cuted}
\usepackage{caption}
\usepackage{soul}
\newacronym{vla}{VLA}{Vision-Language-Action} 
\newacronym{mpc}{MPC}{Model Predictive Control} 
\newacronym{srbd}{SRBD}{Single Rigid Body Dynamics}
\newacronym{grf}{GRF}{ground reaction force} 
\newacronym{com}{CoM}{center of mass} 
\newacronym{ud}{UD}{underdamped} 
\newacronym{cd}{CD}{critically damped} 
\newacronym{od}{OD}{overdamped} 
\newacronym{fft}{FFT}{Fast Fourier Transform}
\newacronym{rms}{RMS}{root-mean-square}
\newacronym{armpc}{ARMPC}{arm-aware MPC}
\newacronym{VLA}{VLA}{Vision-language-action}

\AtBeginDocument{
    \setlength{\abovedisplayskip}{5pt plus 2pt minus 2pt}
    \setlength{\belowdisplayskip}{5pt plus 2pt minus 2pt}
    \setlength{\abovedisplayshortskip}{2pt plus 1pt}
    \setlength{\belowdisplayshortskip}{3pt plus 1pt minus 1pt}
}

\hypersetup{hidelinks}

\title{\LARGE \bfseries LocoPilot: Single-Operator Bilateral Teleoperation \\
for Loco-Manipulation Policy Learning}

\author{
Antonio Langella$^{*,1}$,
Giovanni Minelli$^{*,1}$,
Giovanni B. Dessy$^{*,1}$,
Lorenzo Amatucci$^{1}$, \\
Giulio Turrisi$^{1}$,
Victor Barasuol$^{1}$,
and Claudio Semini$^{1}$%
\thanks{$^{1}$Dynamic Legged Systems Lab, Istituto Italiano di Tecnologia (IIT), Italy. Email: name.lastname@iit.it.
$^{*}$Equal contribution.}%
}

\begin{document}
\bstctlcite{BSTcontrol}

\maketitle

\thispagestyle{empty}
\pagestyle{empty}


\input{0-Abstract}

\input{1-Introduction}

\input{2-Background}

\input{3-Method}

\input{4-Results}

\input{5-Limitations}

\input{6-Conclusions}







\bibliographystyle{IEEEtran}
\bibliography{literature.bib}

\end{document}

%% file: 0-Abstract.tex
\begin{strip}
\vspace{-1.3cm}

    \centering
    \includegraphics[width=0.8\textwidth]{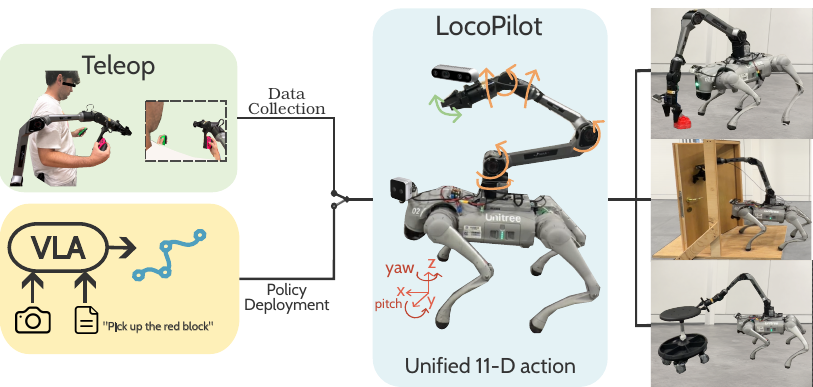}
\captionof{figure}{Overview of the proposed LocoPilot framework.
Upper left: the single-operator teleoperation interface, in which a
leader arm and two handheld controllers command the manipulator,
gripper, base motion, and body posture during data collection.
Lower left: autonomous deployment, where a Vision-Language-Action (VLA)
policy conditioned on visual observations and a language instruction
replaces the operator. Both command the arm-equipped quadruped (center)
through the same loco-manipulation action interface. Right:
representative loco-manipulation tasks executed with the platform. \textbf{Code available at:} \url{https://github.com/iit-DLSLab/LocoPilot}}
    \label{fig:system_overview}
\end{strip}

\begin{abstract}
Learning loco-manipulation policies requires demonstrations that
capture coordinated base and arm motion on the target embodiment.
Robot-free interfaces allow demonstrations to be collected
independently of the target robot, increasing flexibility but
without directly capturing platform-specific commands, proprioception, and
body-mounted observations. On-robot teleoperation allows to capture these signals directly, but requires an interface capable of commanding the complete system concurrently.
This paper presents LocoPilot, a single-operator bilateral teleoperation and
data-collection framework implemented on a 12-DoF Unitree Go2
quadruped equipped with a 6-DoF AgileX Piper L manipulator, an Intel
RealSense D435 wrist camera, and an Intel RealSense D405 head
camera.  A second 6-DoF Piper L arm, used as the leader and two
handheld controllers provide simultaneous control of the base,
body posture, manipulator, and gripper. The follower arm interaction forces are estimated without a
dedicated force/torque sensor and rendered to the operator through a
passivity-constrained haptic channel. A ROS~2 pipeline records
synchronized multimodal observations together with a unified
11D action shared by teleoperation and
policy deployment. Compared to prior work, LocoPilot uniquely unifies robot-in-the-loop legged data collection,
force feedback without motion-capture instrumentation, and the
same action interface at deployment. We evaluate the interface across multiple
loco-manipulation tasks and validate the complete
collection-to-deployment pipeline by fine-tuning MolmoAct2 \cite{MOLMO} on
pick-and-place demonstrations and deploying on
the physical robot without action retargeting or additional inverse
kinematics.
\end{abstract}

%% file: 1-Introduction.tex
\section{Introduction}
\label{sec:introduction}

Robots capable of both locomotion and manipulation extend physical interaction beyond the workspace of fixed-base manipulators. Arm-equipped quadruped robots are particularly suited to unstructured environments, where they combine legged mobility with physical interaction capabilities. Prior work has explored legged loco-manipulation through model-based whole-body control \cite{8794273}, or learned
coordination of locomotion and manipulation \cite{fu2022deep,RobotDuet}.
More recently, generalist
\gls{vla} models have gained increasing attention
for learning manipulation behaviors from multimodal demonstrations,
leveraging large-scale visual and language representations \cite{open_x_embodiment_rt_x_2023,kim24openvla,Human2LocoMan,MoManipVLA}. For loco-manipulation, however, such
demonstrations must capture both base and arm actions on the target
embodiment. Ideally, the collection interface should therefore
minimize the collection--deployment gap while preserving direct
operator authority over all task-relevant commands. Existing systems trade off these requirements in different ways:

\textbf{Teleoperation systems} place a human directly in the control loop of the physical robot, combining the operator task understanding with the constraints of the target embodiment. Leader-follower interfaces with direct joint correspondence provide intuitive control of fixed-base manipulators \cite{GELLO,Team2024ALOHA2A}, while mounting the same kind of interface on a wheeled base extends the concept to mobile manipulation, recording base motion alongside bimanual manipulation \cite{fu2024mobile}. XR-based systems provide more general cross-embodiment interfaces
\cite{cheng2024tv,zhao2025xrobotoolkit}, and DROID demonstrates the use of teleoperation for standardized large-scale data collection \cite{DROID}.
Immersive interfaces additionally stream the robot's own onboard imagery to the operator~\cite{cheng2024tv}, allowing real-time inspection of the visual inputs that will directly condition the downstream policy.
For arm-equipped quadrupeds, previous systems have used wearable motion capture, joysticks, haptic devices, or hand gestures \cite{TeleMAN,SpaceBert,RobotTeleoperativo,quadrupedgesture, xin2020bounded}, and XR headsets have been used to collect whole-body demonstrations for policy learning \cite{qiu2025wildlma,Human2LocoMan}. Collecting directly on the target robot reduces observation and action mismatch, but simultaneously commanding all task-relevant DoFs becomes increasingly difficult for a single operator.

\textbf{Robot-free interfaces} make the opposite trade-off. The Universal Manipulation Interface (UMI) \cite{chi2024universal} records demonstrations using a sensorized handheld gripper, allowing the operator to demonstrate manipulation without the kinematic constraints of the target robot. This improves collection flexibility, but omits embodiment-specific signals such as base
commands, proprioception, and body-mounted observations. Deployment therefore requires an additional mapping from demonstrated
end-effector motion to the target embodiment, as in UMI on Legs \cite{UMIlegs2024}.
This work aims to combine the collection-deployment consistency
of on-robot teleoperation with the flexibility of learning-based
interfaces for arm-equipped quadrupeds. As illustrated in
Fig.~\ref{fig:system_overview}, both the human operator and the
learned policy act through the same loco-manipulation interface,
while the resulting demonstrations are collected directly on the
target platform. Achieving this requires addressing two main
challenges. The first challenge is enabling a single operator to command the
entire loco-manipulation action concurrently. Reaching and grasping
may require the robot to translate, rotate, or change body posture
while the arm is moving, so sequential base and manipulator control
produces segmented demonstrations rather than the smooth continuous
coordination desired from the final policy. Existing systems address
this through multiple operators, wearable instrumentation, or
partial autonomy, introducing additional coordination or removing
task decisions from the demonstrator.

To retain these decisions under single-operator control, we propose a bimanual interface design where a handheld controller is mounted on the end-effector handle of the leader arm. The dominant hand articulates the manipulator while
operating the mounted controller, and the non-dominant hand provides
the remaining base commands through a second controller. Together,
the two interfaces provide concurrent control of planar base
velocity, yaw rate, body pitch, arm configuration, and gripper
aperture. Interaction forces at the follower are additionally
estimated online and rendered through the leader arm, providing
contact feedback without a dedicated force/torque sensor.

The second challenge is maintaining consistency between collection
and deployment. On a legged manipulator, high-level commands are
executed through a locomotion and manipulation stack whose behavior
also shapes the recorded motion. Replacing this stack at deployment
can therefore shift the distribution of observations and executed
motions away from the demonstrated one, even when the high-level
action representation is unchanged.
We avoid this by using the same 11-dimensional loco-manipulation
action in both regimes, generated by the operator during collection
and by the learned policy at deployment, and by keeping the same
downstream execution modules: the locomotion policy and leg controller,
the arm controller, and the gripper interface.
Table~\ref{tab:comparison} compares LocoPilot with the closest prior
systems in terms of the teleoperation and data-collection properties
discussed above.
The contributions of this work can be summarized as follows:
\begin{itemize}
    \item A single-operator bilateral teleoperation interface for
    arm-equipped quadrupeds, combining concurrent base, body, arm,
    and gripper control with sensorless force reflection and an
    operator-side passivity layer.

    \item A unified 11-dimensional action interface shared between
    teleoperation and learned-policy deployment, with the same
    locomotion and manipulation controllers used in both regimes.

    \item A synchronized learning-oriented data pipeline recording
    multi-camera RGB-D observations, robot state, odometry, and
    commanded actions, validated through multiple teleoperation
    tasks and real-robot deployment of a fine-tuned MolmoAct2 policy.
\end{itemize}

\begin{table}[t]
\centering
\caption{Comparison with the closest systems.}
\label{tab:comparison}
\setlength{\tabcolsep}{1.5pt}
\footnotesize
\begin{tabular}{@{}>{\raggedright\arraybackslash}p{2.7cm}cccccc@{}}
\hline
& \scriptsize\textbf{M.ALOHA} & \scriptsize\textbf{TeLeMan} & \scriptsize\textbf{UMI-L.} & \scriptsize\textbf{WildLMa} & \scriptsize\textbf{H2LM} & \scriptsize\textbf{Ours} \\
& \cite{fu2024mobile} & \cite{TeleMAN} & \cite{UMIlegs2024} & \cite{qiu2025wildlma} & \cite{Human2LocoMan} & \\
\hline
Legged embodiment                & --         & \checkmark & \checkmark & \checkmark & \checkmark & \checkmark \\
Robot in the loop                & \checkmark & \checkmark & --         & \checkmark & \checkmark & \checkmark \\
Base action + proprioception     & \checkmark & \checkmark & --         & \checkmark & \checkmark & \checkmark \\
Body-mounted view                & \checkmark & \checkmark & --         & \checkmark & \checkmark & \checkmark \\
Robot-operator force feedback    & --         & --         & --         & --         & --         & \checkmark \\
Direct leader–follower joint mapping         & \checkmark & --         & -- & -- & -- & \checkmark \\
Same action interface at deploy  & \checkmark & n/a        & --         & \checkmark & \checkmark & \checkmark \\
\hline
\end{tabular}
\end{table}


%% file: 2-Background.tex
\section{System Overview}
\label{sec:SystemOverview}
In this section we provide an overview of the hardware and software components that compose our framework.

\subsection{Hardware}
\label{sec:hardware}
Figure \ref{fig:system_overview} shows both the teleoperation setup and the system being controlled.
In particular, the follower platform, shown on the right, consists of a 12-DoF Unitree Go2 quadruped equipped with a 6-DoF AgileX Piper L manipulator and parallel gripper.
Two RGB-D cameras provide complementary observations: an Intel RealSense D435 mounted near the gripper captures close-range manipulation, while an Intel RealSense D405 mounted on the quadruped head, thanks to its larger field of view, observes the wider workspace in front of the robot.
This platform is the same during both demonstration collection and autonomous deployment.

\begin{figure}[!t]
    \centering
    \includegraphics[width=\linewidth]{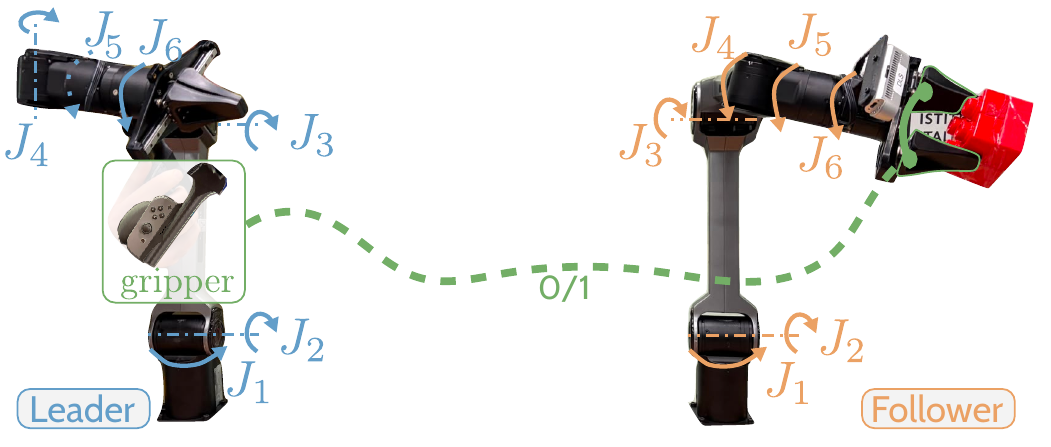}
    \caption{The two sides of the manipulation system setup. Left: the leader station,
    with the end-effector handle, and the handheld controller mounted. Right:
    the follower platform, a Piper L arm and a gripper-mounted
    RGB-D camera.}
    \label{fig:two_arm_teleop_station}
\end{figure}

The leader station, on the left of Fig.~\ref{fig:two_arm_teleop_station}, is used only during data collection.
It is built around a second Piper L arm serving as a leader device.
A custom handle at the end-effector improves ergonomics and holds one of two Nintendo Joy-Con controllers~\cite{nintendoJoyCon}.
The operator grips this handle with the dominant hand and articulates the leader arm. With the thumb of the same hand, the operator reaches the controller's analog stick and buttons, which respectively control the quadruped robot pitch and yaw rate and the episode markers, while the index finger controls the gripper through the trigger button.
With the non-dominant hand the operator holds the second controller, which commands the base velocity.
The configuration shown in Fig.~\ref{fig:two_arm_teleop_station} assumes a right-handed operator: for left-handed operation, a specular handle can be employed while swapping the controller assignment without changing the teleoperation scheme.
To ensure visual consistency, the operator drives the robot using the exact onboard camera streams recorded for downstream policy training, displayed in real time at the leader station. This visual alignment ensures the demonstrator and the policy share the exact same perspective. At the same time, placing all commands under a single operator makes it easy to execute base and arm movements concurrently instead of sequentially; for instance, the operator can steer and turn the quadruped while reaching toward an out-of-reach object in a single coordinated motion.

\subsection{Software Architecture}
\label{sec:software}

\begin{figure*}[ht]
    \centering
    \includegraphics[width=0.9\linewidth]{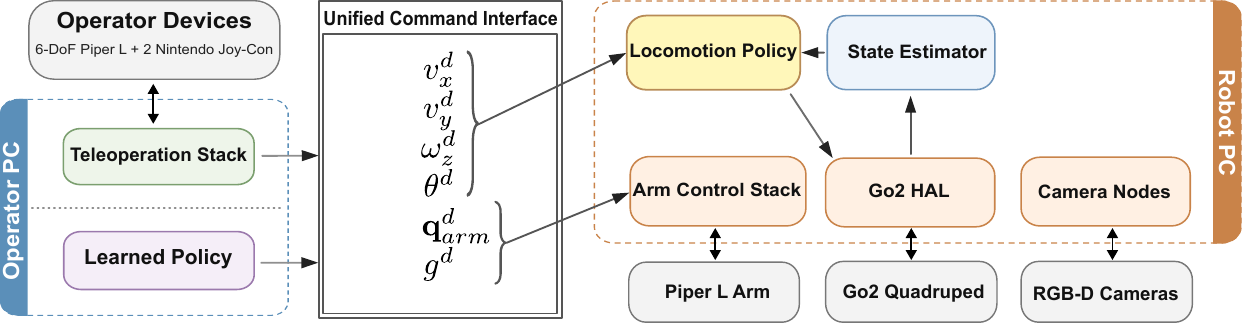}
    \caption{Software modules and their placement. The robot-side
    modules are identical during demonstration collection and
    autonomous deployment; only the operator-side stack changes, with
    the leader arm, teleoperation loop, and recorder replaced by a
    single policy-inference node.}
    \label{fig:architecture}
    \vspace{-0.3cm}
\end{figure*}

On the software side, the framework is composed of interconnected, distributed modules running on the quadruped's onboard computer and an operator workstation, as shown in Fig.~\ref{fig:architecture}.
In particular, the quadruped's onboard computer runs: the reinforcement-learning base policy of~\cite{turrisi26BinWalker}, which converts base-velocity and body-pitch references into joint position targets for a low-level PD controller; the MUSE state estimator~\cite{MUSE}, which provides the state estimates used as input by the policy; the arm control stack; and the camera nodes, which publish the two RGB-D streams.

As a direct consequence of unifying the action and observation interfaces, the same software modules run on the robot both for data collection and for autonomous deployment.
What changes between the two modes is the operator side.
During demonstration collection the workstation runs: the leader arm control stack; the teleoperation module described in Sec.~\ref{sec:leader-follower} that couples leader and follower and renders contact back to the operator; and the recording module that assembles episodes from the observation, state, and command streams.
During autonomous deployment a single policy-inference node replaces all three, receiving the same observations and emitting the same commands the teleoperation loop previously produced.
The switch between the operator and the policy therefore falls at one interface, and everything after it is unaffected by which of the two is in control.

Having defined explicit software interfaces that decouple the teleoperation and autonomous deployment stack from the underlying robot hardware, each hardware-dependent subsystem can be replaced independently, provided that it preserves the corresponding input-output contract.
In particular, the locomotion subsystem, including the base, leg controller, reinforcement-learning policy, and state estimator, can be replaced by any implementation that accepts commanded planar velocity and absolute body pitch and exposes the leg states and base-velocity estimate required by the higher-level stack.
Likewise, the manipulation subsystem can be replaced by any arm and
controller implementation that supports joint-space impedance control
and exposes joint position, velocity, and torque feedback.
All components above these interfaces, including the teleoperation logic, action representation, and data-collection pipeline, remain unchanged, which reduces the engineering effort required to port the framework to a different platform.
The present implementation uses ROS~2 Jazzy with Cyclone DDS.

%% file: 3-Method.tex
\section{Teleoperation and Data-Collection Framework}
\label{sec:teleop-brackground}

\subsection{Leader-Follower Arm Mapping}
\label{sec:leader-follower}
A haptic bilateral teleoperation scheme is used to map the leader arm motion to the follower while reflecting interaction forces back to the operator, as shown in Fig.~\ref{fig:two_arm_teleop_station}.
Because leader and follower share the same kinematic structure, the coupling between them can be imposed directly in joint space: the follower directly tracks the leader configuration, and the latter renders the joint-space interaction torques experienced by the follower.
We model both manipulators with the standard joint-space rigid-body
dynamics
\begin{equation}
\left\{
\begin{aligned}
\mathbf{M}_{l}\ddot{\mathbf{q}}_{l}
+\mathbf{C}_{l}\dot{\mathbf{q}}_{l}
+\mathbf{g}_{l}+\mathbf{f}_{l}
&=\boldsymbol{\tau}_{l}+\boldsymbol{\tau}_{h},\\
\mathbf{M}_{f}\ddot{\mathbf{q}}_{f}
+\mathbf{C}_{f}\dot{\mathbf{q}}_{f}
+\mathbf{g}_{f}+\mathbf{f}_{f}
&=\boldsymbol{\tau}_{f}+\boldsymbol{\tau}_{\mathrm{ext}} .
\end{aligned}
\right.
\label{eq:arm-dynamics}
\end{equation}
where subscripts $l$ and $f$ denote leader and follower quantities, respectively.
In the usual notation, $\mathbf{q}_{*}$ is the joint configuration, $\mathbf{M}_{*}$ the inertia matrix, $\mathbf{C}_{*}$ the Coriolis/centrifugal matrix, and $\mathbf{g}_{*}$ the gravity torque.
The term $\mathbf{f}_{*}$ denotes the friction, while $\boldsymbol{\tau}_{*}$ is the actuator torque.
Finally, $\boldsymbol{\tau}_{h}$ and $\boldsymbol{\tau}_{\mathrm{ext}}$ are the human operator and environment
interaction torques.
The two models are identified separately, starting from their nominal values, since the leader carries the operator handle while the follower carries the gripper and the wrist camera.
As usual, $\mathbf{M}_{*}$ is symmetric positive definite and $\dot{\mathbf{M}}_{*}-2\mathbf{C}_{*}$ is skew-symmetric.
Throughout, the dependence on time is omitted where unambiguous, and made explicit only for quantities evaluated at a specific instant.

\subsubsection{Position channel}

The follower tracks the leader through a local PD law with gravity compensation,
\begin{equation}
\boldsymbol{\tau}_{f}^{d} =
-\mathbf{K}_{p}\left(\mathbf{q}_{f}-\mathbf{q}_{l}\right)
-\mathbf{K}_{d}\left(\dot{\mathbf{q}}_{f}-\dot{\mathbf{q}}_{l}\right)
+\mathbf{g}_{f}(\mathbf{q}_{f}),
\label{eq:follower-law}
\end{equation}
with $\mathbf{K}_{p},\mathbf{K}_{d}\in\mathbb{R}^{6\times 6}$ diagonal and positive definite.
The superscript $d$ denotes a desired quantity; we assume the drives realize $\boldsymbol{\tau}_{f}^{d}$ as the actuator torque $\boldsymbol{\tau}_{f}$ of~\eqref{eq:arm-dynamics} without saturation.

\subsubsection{Force channel}

The leader receives an estimate of the external joint torque induced by the interaction between the follower and the environment.
The estimate is obtained using a generalized-momentum residual filter~\cite{deluca2005sensorless} on the identified follower model, which does not require dedicated force/torque sensors.
Let $\mathbf{p}=\mathbf{M}_{f}(\mathbf{q}_{f})\dot{\mathbf{q}}_{f}$ be the generalized momentum of the follower, $\mathbf{p}(0)$ its value at the instant $t=0$ at which the observer is initialized, and $\boldsymbol{\beta}=\mathbf{g}_{f}+\mathbf{f}_{f}-\mathbf{C}_{f}^{\top}\dot{\mathbf{q}}_{f}$ the sum of its gravity, friction and Coriolis terms.
Differentiating $\mathbf{p}$ and using the skew-symmetry of $\dot{\mathbf{M}}_{f}-2\mathbf{C}_{f}$ gives $\dot{\mathbf{p}}=\boldsymbol{\tau}_{f}-\boldsymbol{\beta}+\boldsymbol{\tau}_{\mathrm{ext}}$, so the momentum change $\mathbf{p}-\mathbf{p}(0)$ accumulated since initialization differs from the change predicted by the known torques, the integral term below, exactly by the contribution of $\boldsymbol{\tau}_{\mathrm{ext}}$.
The residual evaluates this mismatch and feeds it back through $\mathbf{K}_{I}$,
\begin{equation}
\mathbf{r} = \mathbf{K}_{I}\!\left[\mathbf{p}-\mathbf{p}(0)
-\int_{0}^{t}\!\left(\boldsymbol{\tau}_{f}^{m}-\boldsymbol{\beta}
+\mathbf{r}\right)\mathrm{d}s\right],
\label{eq:residual}
\end{equation}
where the actuator torque is replaced by the current-derived joint-effort feedback $\boldsymbol{\tau}_{f}^{m}\approx\boldsymbol{\tau}_{f}$ reported by the follower drives.
The observer is initialized at teleoperation engagement, when the follower is at rest and $\mathbf{p}(0)=\mathbf{0}$.
Under the rigid-body dynamics of ~\eqref{eq:arm-dynamics}, the residual obeys $\dot{\mathbf{r}}=\mathbf{K}_{I}(\boldsymbol{\tau}_{\mathrm{ext}}-\mathbf{r})$, so $\mathbf{r}$ estimates the external joint torque to first order with bandwidth $\mathbf{K}_{I}$ and without requiring joint acceleration.
To suppress high-frequency residual components caused by measurement noise and unmodeled dynamics, we apply a low-pass filter on the residual at $f_{c}=\SI{5}{\hertz}$; in the equations below, $\mathbf{r}$ denotes the filtered residual.

On the leader side, gravity compensation and light damping are combined with the reflected estimate,
\begin{equation}
\boldsymbol{\tau}_{l} =
\mathbf{g}_{l}(\mathbf{q}_{l})
-\mathbf{D}_{l}\dot{\mathbf{q}}_{l}
+\alpha\,\boldsymbol{\Lambda}\,\mathbf{r},
\label{eq:leader-law}
\end{equation}
where $\boldsymbol{\Lambda}$ is a diagonal reflection gain, $\mathbf{D}_{l}\succ 0$ a diagonal damping matrix, and $\alpha\in[0,1]$ the modulation supplied by the passivity layer (described below).
No position term appears in ~\eqref{eq:leader-law}: the leader is free to move under damping alone, and the operator feels the follower only when the latter interacts with the environment.

The force channel is therefore designed to convey informative interaction cues rather than reproduce the environment impedance exactly.
As shown in~\cite{wildenbeest2013haptic}, even low-frequency force feedback, as provided by our filter at $f_c$, can deliver most of the benefits in task completion time and control effort.
In learning from demonstration, haptic feedback during teleoperated data collection has also been shown to improve demonstration throughput and the performance of the resulting autonomous policies~\cite{cuan2024haptic}.

\subsubsection{Passivity layer}
\label{subsec:passivity-layer}
Position--force coupling does not, in general, make the teleoperator passive by construction~\cite{lawrence1993stability}.
Classic bilateral frameworks constrain both sides of the chain~\cite{franken2011bilateral}; we instead passivize only the operator-facing port, a single-port approach with precedent in Time-Domain Passivity Control~\cite{hannaford2002time}, which leaves the follower command unmodified and therefore identical to the one used at deployment.
Following~\cite{ferraguti2013tank}, the active reflection $\boldsymbol{\tau}_{r}^{d}=\boldsymbol{\Lambda}\mathbf{r}$ is budgeted by an energy tank $H$, filled by the power dissipated by the leader damping and drained by the power the reflection injects,
\begin{equation}
\dot{H} = \dot{\mathbf{q}}_{l}^{\top}\mathbf{D}_{l}\dot{\mathbf{q}}_{l}
-\alpha\,\boldsymbol{\tau}_{r}^{d\top}\dot{\mathbf{q}}_{l},
\qquad 0<H(0)=H_{0}\leq H_{\max},
\label{eq:tank}
\end{equation}
with $H$ upper-bounded by $H_{\max}$ so that no unbounded reservoir accumulates, and initialized at the same instant $t=0$ as the residual.
Writing $P_{k}^{d}=\boldsymbol{\tau}_{r,k}^{d\top}\dot{\mathbf{q}}_{l,k}$ for the power the reflection would inject at control sample $k$, the modulation is
\begin{equation}
\alpha_{k} =
\begin{cases}
1, & P_{k}^{d}\leq 0,\\[4pt]
\displaystyle\min\!\left(1,\max\!\left(0,\frac{H_{k}}{T_\delta\,P_{k}^{d}}\right)\right), & P_{k}^{d}>0,
\end{cases}
\label{eq:modulation}
\end{equation}
with $T_\delta$ the control period.
Reflection passes unchanged whenever it absorbs energy from the operator, and is scaled down when it would inject more than the tank holds over the next interval.
Taking the storage $S=\tfrac{1}{2}\dot{\mathbf{q}}_{l}^{\top}\mathbf{M}_{l}\dot{\mathbf{q}}_{l}+H$, and using the skew-symmetry of $\dot{\mathbf{M}}_{l}-2\mathbf{C}_{l}$ together with Eqs.~\eqref{eq:leader-law} and~\eqref{eq:tank}, the damping and reflected-power terms cancel and $\dot{S}=\boldsymbol{\tau}_{h}^{\top}\dot{\mathbf{q}}_{l}-\dot{\mathbf{q}}_{l}^{\top}\mathbf{f}_{l}\leq\boldsymbol{\tau}_{h}^{\top}\dot{\mathbf{q}}_{l}$, since the leader joint friction is dissipative.
The same bound holds when $H_{\max}$ is reached and the excess energy is discarded.
As $S$ is non-negative, the haptic subsystem is passive at the operator port $(\boldsymbol{\tau}_{h},\dot{\mathbf{q}}_{l})$, and the guarantee does not extend to the complete operator--leader--follower--environment chain.
Since the residual enters only through the injected power, errors in the follower model degrade the rendered cue but not the energy accounting.

\subsubsection{Engagement and safety}

Teleoperation is engaged after an automatic procedure which aligns the follower and the leader configurations, preventing a position-error transient at coupling.
The tank is initialized to $H_0$ at engagement, providing an initial energy budget for active reflection.
The operator can suspend coupling at any time, and loss of the leader state stream triggers the same transition, causing the follower to hold its current configuration.
Commanded positions and torques are clamped to the hardware limits.

\subsection{Unified Loco-Manipulation and Learning Interface}
\label{sec:unified-action}

In both regimes of Section~\ref{sec:software}, the robot is commanded through the same 11-dimensional loco-manipulation action, written by the operator during collection and by the policy at deployment,
\begin{equation}
\mathbf{a} =
\begin{bmatrix}
v_x^{d} &
v_y^{d} &
\omega_z^{d} &
\theta^{d} &
\mathbf{q}_{\mathrm{arm}}^{d\top} &
g^{d}
\end{bmatrix}^{\top},
\label{eq:unified-action}
\end{equation}
where $v_x^{d}$ and $v_y^{d}$ are the desired linear velocities of the base in the horizontal plane, expressed in a gravity-aligned frame that follows the base yaw, $\omega_z^{d}$ the desired yaw rate, $\theta^{d}$ the desired body pitch, $\mathbf{q}_{\mathrm{arm}}^{d}\in\mathbb{R}^{6}$ the follower-arm joint references, and $g^{d}$ the gripper aperture.
The command is divided among three low-level interfaces: the base velocities and pitch references are sent to the locomotion policy, the arm references to the manipulator controller, and the gripper reference to the gripper interface.
During deployment, the joint-position references produced by the policy are temporally interpolated by the arm trajectory interface, which samples the resulting trajectory to provide both position and velocity references to the low-level joint controller of ~\eqref{eq:follower-law}.
During demonstration collection, these references are provided directly by the leader position and velocity, while the remaining action entries come from the two handheld controllers.

The arm command $\mathbf{q}_{\mathrm{arm}}^{d}$ is expressed in joint space rather than as an end-effector pose~\cite{GELLO,fu2024mobile}.
Since leader and follower share the same kinematics, the recorded action is the reference that the arm controller receives, and deployment requires no inverse kinematics, which would have to select among solution branches and becomes ill-conditioned near the extended configurations used for reaching.
We note that this choice makes recorded arm actions specific to the follower's kinematics, preventing them from transferring directly to a different manipulator, although forward kinematics maps each of them to a unique end-effector pose, so that a task-space version of the dataset can be derived offline when needed.

\subsubsection{Locomotion policy}

The base velocities and pitch reference are tracked by a reinforcement-learning policy, adapted from BinWalker~\cite{turrisi26BinWalker}, that outputs the 12 desired leg joint positions passed to the quadruped low-level controller.
Since the arm changes the mass distribution of the platform and transmits a wrench to its base, the measured arm configuration is part of the proprioceptive observation,
\begin{equation}
\mathbf{o} =
\left(
\mathbf{v},\;
\boldsymbol{\omega},\;
\mathbf{g}_{b},\;
\mathbf{q}_{\mathrm{leg}},\;
\dot{\mathbf{q}}_{\mathrm{leg}},\;
\mathbf{q}_{\mathrm{arm}},\;
v_x^{d},\, v_y^{d},\, \omega_z^{d},\, \theta^{d}
\right),
\label{eq:locomotion-observation}
\end{equation}
alongside the estimated base linear and angular velocities
$\mathbf{v}$ and $\boldsymbol{\omega}$, the gravity direction
$\mathbf{g}_{b}=\mathbf{R}_{b}^{\top}\hat{\mathbf{g}}$ expressed
in the base frame, where $\mathbf{R}_{b}$ maps base-frame vectors to
the world frame, the leg joint positions and velocities, and the
four commanded entries of ~\eqref{eq:unified-action} that the
policy is asked to track. The term $\mathbf{g}_{b}$ encodes the
base roll and pitch and not the yaw, which matches the command
convention of ~\eqref{eq:unified-action}: the pitch reference
$\theta^{d}$ is absolute, whereas the yaw is commanded as a rate. The five
most recent observations are stacked and given to the policy at
each step. On hardware $\mathbf{v}$, $\boldsymbol{\omega}$, and the
base orientation $\mathbf{R}_{b}$ are supplied by the MUSE state
estimator~\cite{MUSE}, in both modes. We also extend the arm-configuration 
randomization of BinWalker's policy at training time to cover the 
workspace required by this platform. Specifically, this accounts 
for the forward-extended configurations that occur when reaching 
for and transporting objects.
The policy therefore adapts the leg motion to the current arm configuration, while never commanding the arm itself.
Locomotion and manipulation remain modular, and the policy accounts for their physical coupling through the observation of ~\eqref{eq:locomotion-observation}.

\subsubsection{Policy learning interface}

The recording module is designed so that demonstrations can be used
for policy learning without redefining the robot interface. At each
step, it pairs the synchronized camera images and proprioceptive state
with the command issued by the operator at that instant. Complete
episodes are exported to the LeRobot dataset format~\cite{Cadne2026LeRobotAO}
and can be used for training without further conversion.
Two choices keep the exported data model-agnostic while preserving the
collection--deployment interface. First, the recorded action is the
command received by the low-level controllers, rather than the robot
response measured afterwards. A policy trained on these labels therefore
emits quantities that the same controllers already accept at deployment.
Second, each episode includes colour and depth images from both cameras, 
the proprioceptive state, the action, and custom language instructions, 
allowing a candidate model to condition on any subset of these observations.
The only requirement imposed by the framework is that the model transmits the
loco-manipulation action of ~\eqref{eq:unified-action}, or a subset of
it.
Its output rate is not constrained to match the low-level controllers, since the arm trajectory interface interpolates the joint references to the controller rate, and the remaining entries are held between updates.

%% file: 4-Results.tex
\section{Experimental Evaluation}
\label{sec:Experimental_results}

\subsection{Arm-to-arm Teleoperation}

We evaluate the leader--follower teleoperation interface of Sec.~\ref{sec:leader-follower} on a tabletop pick-and-place task containing both free-space motion and sustained object interaction.
The experiment verifies that the two channels required for demonstration
collection operate simultaneously: the follower tracks the leader motion, while
contact with the environment produces an interaction estimate suitable for haptic
rendering.
To obtain a ground-truth reference for the force channel, the manipulated
object is a known payload of \SI{1.5}{\kilo\gram}, so that during transport the expected
vertical component of the estimated interaction force is known.

\begin{figure}[t]
    \centering
    \includegraphics[width=0.95\linewidth]{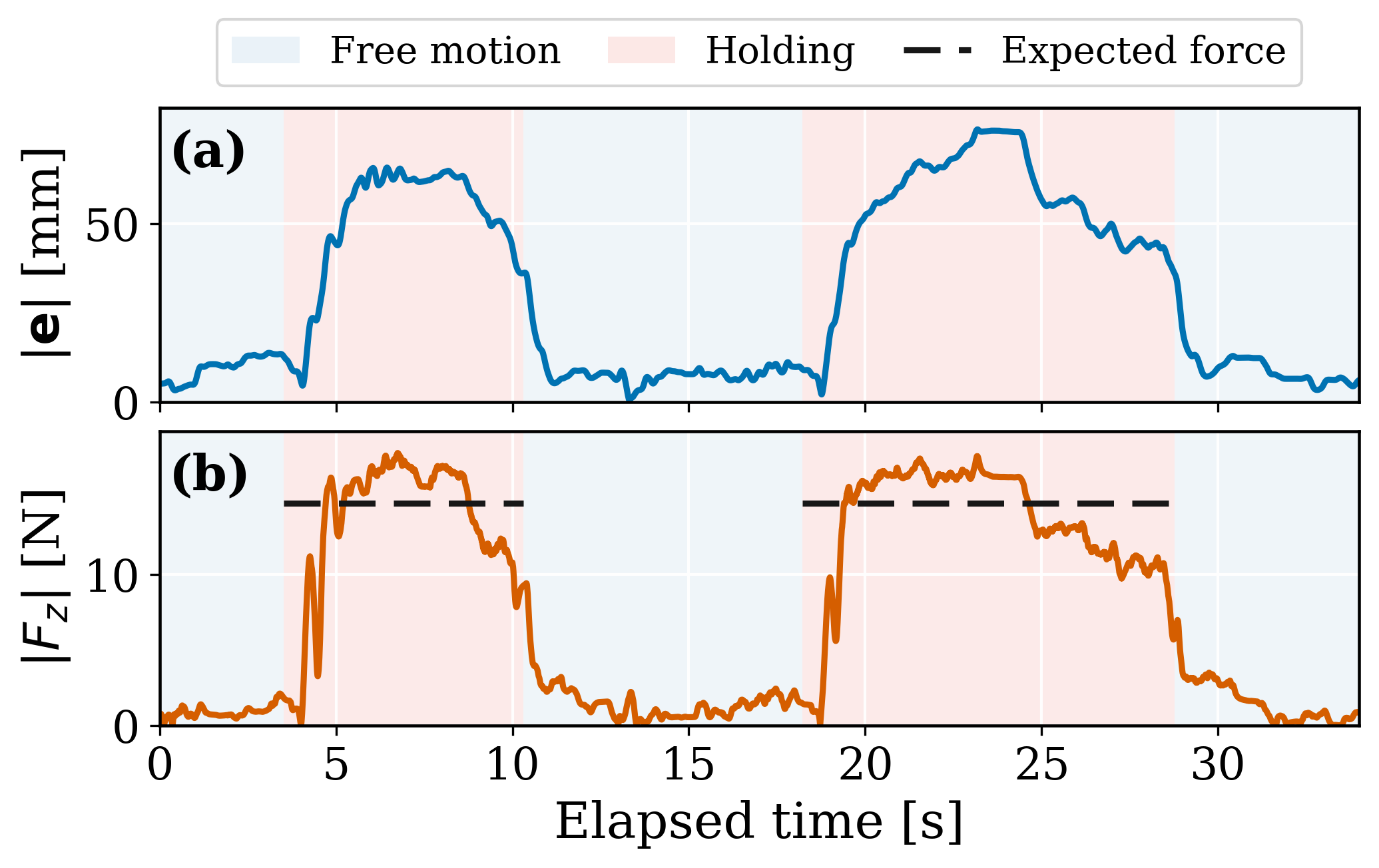}
    \caption{Leader--follower tracking and interaction during a tabletop pick-and-place episode with a payload of known weight. (a) Norm of the end-effector position error $\|\mathbf{e}\|$. (b) Vertical component $|F_z|$ of the estimated interaction force, with the payload weight drawn as a reference line. Blue and red shaded regions mark the free-motion and holding phases respectively.}
    \label{fig:coupling-error}
\end{figure}

The position channel is evaluated through the Cartesian tracking error $\mathbf{e}$ between the leader and follower end-effector positions, while the force channel is characterized by the norm of the estimated interaction force $\mathbf{F}$.
Figure~\ref{fig:coupling-error} reports both over a representative episode, in which the robot first moves in free space, then grasps and transports the payload, and finally releases it.
The weight of the payload is drawn as a reference line, since it is the force the estimate should report while the object is held.

We compare the estimate against the value expected in each phase: zero while the robot moves in free space, and the weight of the payload while it is carried.
Grasping and releasing take a short time, during which the load builds up and vanishes, so the second (\SI{1}{\second}) before and after each of the two transitions is excluded from the comparison.

In free motion the estimated vertical force norm has an RMSE of \SI{1.27}{\newton}, which is the noise floor of the estimate and therefore the smallest interaction the operator can perceive.
While the payload is held, the RMSE against its known weight is \SI{2.25}{\newton}, or \SI{15.3}{\percent} of the reference value.
Tracking error follows the same structure: an RMSE of \SI{8.02}{\milli\metre} in free motion against \SI{60.14}{\milli\metre} during interaction, as expected when environmental constraints oppose the configuration commanded by the leader.
The two signals remain complementary, with the position error quantifying the deviation between commanded and executed motion and the force estimate exposing the interaction responsible for it.

\subsection{Floating-base effects.}
The gravity components of~\eqref{eq:follower-law} and~\eqref{eq:residual} are calculated in the arm base frame using the MUSE-estimated base attitude.
However, the fixed-base model neglects inertial effects
induced by base acceleration and rotation, which the residual
attributes to the environment.
We quantify this contribution over \SI{3.4}{\hour} of recorded loco-manipulation sessions, including the data collection of Sec.~\ref{sec:vla-results}.
At each sample, floating-base inverse dynamics is evaluated
using the recorded arm and base states, both with the estimated
base motion and with the base stationary at the same attitude.
The difference gives the model-predicted torque induced by
base motion, which we pass through the residual dynamics
and the \SI{5}{\hertz} low-pass filter before mapping it to the end-effector as the estimator does. Table~\ref{tab:base-effects} reports the RMS and 95th percentile (p95) of the resulting force norm.
The predicted RMS ranges from \SI{0.60}{\newton} when standing with the
arm static to \SI{2.47}{\newton} during pitching; the corresponding p95
reaches \SI{4.43}{\newton}.

\begin{table}[t]
\centering
\caption{Model-predicted spurious end-effector force due inertial effects induced by base motion.
RMS and p95 are computed from the norm of the predicted inertial
contribution. The fraction is the RMS force divided by the weight of
the arm and gripper, excluding the base link (\SI{3.64}{\kilogram}).}
\label{tab:base-effects}
\setlength{\tabcolsep}{3pt}
\footnotesize
\begin{tabular}{@{}lrrrr@{}}
\hline
\textbf{Base condition} & \textbf{Time} & \textbf{RMS}
& \textbf{p95} & \textbf{RMS / weight} \\
& [min] & [N] & [N] & [\%] \\
\hline
Standing, arm static & 62.9 & 0.60 & 1.37 & 1.7 \\
Standing, arm moving & 45.9 & 0.91 & 2.06 & 2.5 \\
Walking & 24.7 & 1.70 & 3.16 & 4.8 \\
Turning & 15.3 & 1.98 & 3.55 & 5.5 \\
Pitching & 20.2 & 2.47 & 4.43 & 6.9 \\
\hline
\end{tabular}
\end{table}

\begin{figure*}[h!]
\centering
\captionsetup[subfloat]{skip=1pt,captionskip=1pt,farskip=0pt}
\subfloat[Start]{\includegraphics[width=0.32\linewidth]{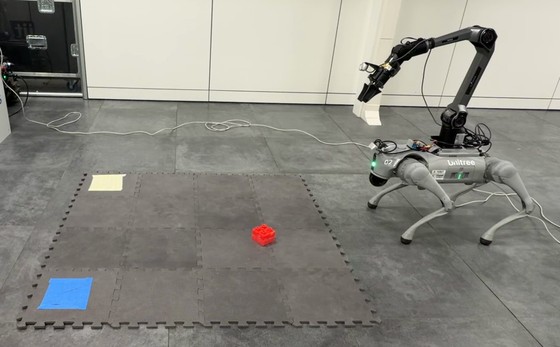}}\hfill
\subfloat[Approach]{\includegraphics[width=0.32\linewidth]{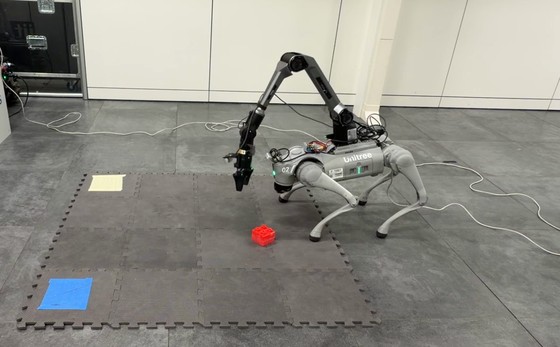}}\hfill
\subfloat[Grasp]{\includegraphics[width=0.32\linewidth]{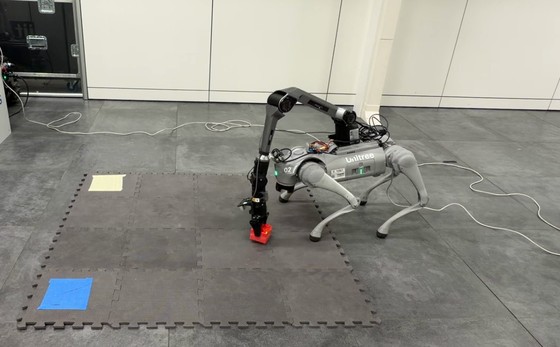}}\\[+0.1ex]
\subfloat[Lift]{\includegraphics[width=0.32\linewidth]{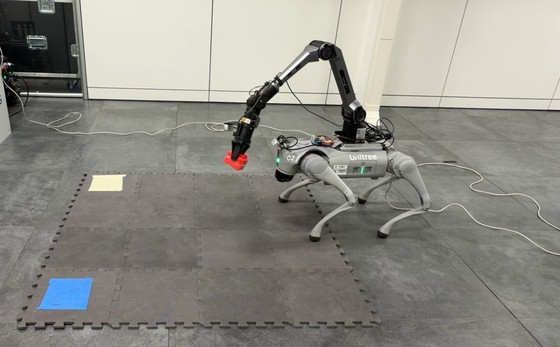}}\hfill
\subfloat[Transport]{\includegraphics[width=0.32\linewidth]{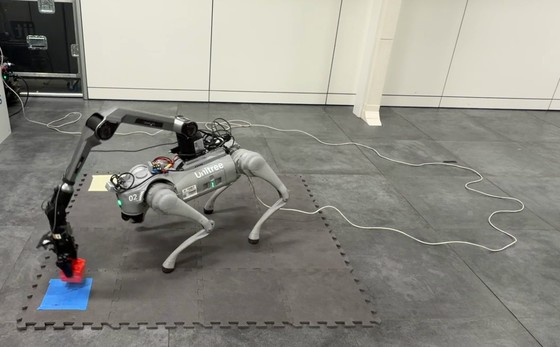}}\hfill
\subfloat[Release]{\includegraphics[width=0.32\linewidth]{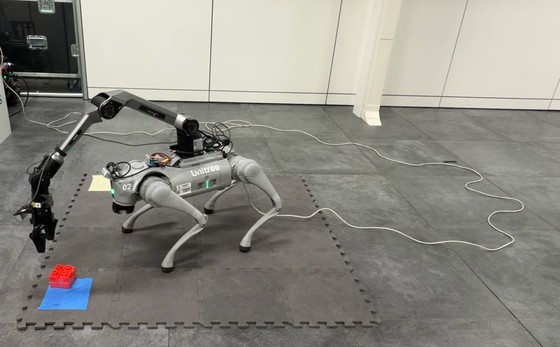}}
\caption{Representative autonomous rollout of the fine-tuned vision-language-action model for the instruction ``Pick the red cube and place it on the blue base'', commanded through the interface of Sec.~\ref{sec:unified-action}. The base approaches the cube (a, b), the body pitches as the gripper closes (c), and the cube is lifted and carried while the base repositions (d, e), then released on the target (f).}
\label{fig:rollout}
\vspace{-0.3cm}
\end{figure*}

\subsection{VLA Deployment}
\label{sec:vla-results}

We evaluate the collected demonstrations by fine-tuning a
vision-language-action model and deploying it on the robot. The
purpose of the experiment is to verify that the recorded episodes are
directly usable for policy learning, and that the coordinated base
and arm motion they contain is reproduced by the learned policy. The
episodes exported by the recorder are used to train the policy, which
then drives the robot through the interface the operator employed,
without action retargeting or inverse kinematics between collection and
deployment.
We instantiate the command interface of
Sec.~\ref{sec:unified-action} with MolmoAct2~\cite{MOLMO}, a vision-language-action model that
predicts action chunks from camera observations and a language
instruction. The choice is motivated by openness: its training data, recipe, and
pretrained checkpoints are released, so the fine-tuning reported here can be repeated in full. In its original paper~\cite{MOLMO} the model is demonstrated only on fixed-base manipulators, so fine-tuning it here also accounts for instructing a new physical embodiment which includes commanding a mobile base.
The dataset collected with the framework covers two tasks, \textit{Grasp}, in which the robot picks up a cube and places it on a colored target, and \textit{Push}, in which it moves the cube to the target without lifting it. In both tasks the target usually lies outside the workspace of the arm at the stance in which the episode starts, so the recorded actions contain coordinated base and arm motion rather than manipulation alone. Either a red or a white cube is used for the collection, and textual instructions are generated from templates over the object and target descriptions, which varies the language input across episodes.
The collection comprises 289 demonstrations, 259 of grasping and 30 of pushing, of which 258 are used for fine-tuning and 31 are held out for validation.
Since the interface allows base and arm to be commanded concurrently, the demonstrations contain the two motions together rather than in sequence: the arm joint references change during \SI{89}{\percent} of the time in which a base velocity or pitch command is active, and \SI{97}{\percent} of the episodes contain at least \SI{1}{\second} of such concurrent motion.
Fine-tuning starts from the released \texttt{MolmoAct2} post-training checkpoint, of \num{5.5}B parameters, and follows the strategies reported by its authors, adapting the vision-language backbone with rank-512 low-rank adapters \cite{hu2022lora} and training the action head in full. The policy runs at \SI{5}{\hertz} and predicts sequences of 10 actions (\SI{2}{\second}), with training and inference running on an RTX 6000 Pro GPU. We select the best checkpoint based on its open-loop performance on held-out episodes, where predicted action chunks are compared against recorded commands, then deploy that checkpoint on the robot.



\textbf{Deployment.} On the robot the policy replaces the operator without any change below the interface: it receives the two camera streams, the proprioceptive state, and a language instruction, and
writes the same command that the teleoperation layer produced during
collection, at the same rate. Example instructions include
``Move the red cube to the blue square'' and
``Pick up the red cube and place it in the blue square.''
Figure \ref{fig:rollout} shows a successful rollout, in which the model drives the base toward the cube, pitches the robot's body to extend the reach of the arm, closes the gripper, and carries the cube to the target region, coordinating base and arm within one continuous motion. The trials are assessed qualitatively, with respect to the ability to solve the task from different starting positions. The policy learns the pick-and-place motion in the simplest scenarios, specifically when both the cube and the target are visible and no complex reorientation is required. Complications arise when the target region leaves the camera's field of view, since the policy must then reorient and navigate based solely on surrounding visual cues, and it sometimes walks and moves the cube in the wrong direction. Another challenging scenario occurs when the base stops too early or too far from the target region, so that the cube is released slightly away from it. In these situations the smooth transition between motions leads the policy to perform corrections, nudging the cube toward the target or re-approaching it for a second grasp until the task is completed.
\vspace{-0.1cm}

%% file: 5-Limitations.tex
\section{Limitations and Applicability}
\label{sec:limitations}

The residual of ~\eqref{eq:residual} is computed from a model of the arm alone.
The floating base is treated as inertial, so whenever it accelerates or tilts the resulting inertial and gravity effects appear at the joints and the observer attributes them to the environment (see Table~\ref{tab:base-effects}).
A second error comes from the identified inertial and friction parameters, whose residual mismatch, dominated by the static friction of the joint drives, biases the estimate as a function of the configuration.
Neither contribution can be removed entirely, and together they set a free-motion floor, which is the smallest interaction the operator can perceive.
The base-induced terms can be computed from the base state already estimated on board, so compensating them in the residual would lower the floor during base motion, and we leave it to future work.

These errors affect what the operator perceives but not the stability of the interface.
The passivity layer of Sec.~\ref{subsec:passivity-layer} accounts for measured power flow, so an inaccurate residual costs force feedback accuracy rather than safety.
The force channel is designed as an extension of the position channel, such that disabling it does not change the recorded action nor the collection pipeline.

We found that the force channel is most useful when the environment constrains the end-effector motion in ways not visible to onboard cameras, and when the resulting interaction is sustained rather than impulsive.
Opening a door or rotating a valve constrains the end effector along a path that the operator has to follow, and pushing or pulling a load requires regulating a force over the whole motion.
In these cases the operator perceives the constraint directly instead of inferring it from the position error.
Tasks dominated by free-space motion, or in which contact is brief and light with respect to the floor reported above, gain correspondingly less from the force channel.

Our evaluation covers a single hardware configuration, training the policy on two task families across a limited number of qualitative trials.
While our interface enables coordinated, simultaneous base and arm movements rather than requiring sequential control, we do not quantitatively evaluate its efficiency relative to sequential teleoperation. We leave this comparison, together with controlled studies on interface control modes, dataset scaling, and contact feedback, to future work.

%% file: 6-Conclusions.tex
\section{CONCLUSIONS}
\label{sec:Conclusion}

We introduced LocoPilot, a single-operator bilateral teleoperation framework for collecting loco-manipulation demonstrations on an arm-equipped quadruped. Its unified 11-dimensional action interface enables simultaneous base and arm control and direct reuse of recorded commands for policy learning and deployment, without action retargeting. We demonstrated the complete pipeline by fine-tuning MolmoAct2 and deploying it for autonomous mobile manipulation. The framework provides a foundation for scalable collection of embodiment-specific demonstrations and future loco-manipulation policy learning.